\documentclass[conference]{IEEEtran}
\IEEEoverridecommandlockouts
\usepackage{dutchcal}
\usepackage{cite}
\usepackage{amsmath,amssymb,amsfonts}
\usepackage{algorithmic}
\usepackage{algorithm}
\usepackage{mathtools}
\usepackage{graphicx}
\usepackage{textcomp}
\usepackage{xcolor}
\usepackage{adjustbox}
\usepackage{xspace}
\usepackage{multirow}
\usepackage{authblk}
\usepackage{colortbl}
\usepackage{bbding}
\usepackage{adjustbox}
\usepackage{hyperref}
\usepackage{booktabs}

\def\BibTeX{{\rm B\kern-.05em{\sc i\kern-.025em b}\kern-.08em
    T\kern-.1667em\lower.7ex\hbox{E}\kern-.125emX}}

\makeatletter
\DeclareRobustCommand\onedot{\futurelet\@let@token\@onedot}
\def\@onedot{\ifx\@let@token.\else.\null\fi\xspace}

\def\eg{\emph{e.g}\onedot} 
\def\ie{\emph{i.e}\onedot}

\def\etal{\emph{et al}\onedot}

\begin{document}

\title{UIE-UnFold: Deep Unfolding Network with Color Priors and Vision Transformer for Underwater Image Enhancement}

\author[1*]{Yingtie Lei\thanks{* These authors contribute equally to this work.}}
\author[2*]{Jia Yu}
\author[3]{Yihang Dong}
\author[3]{Changwei Gong}
\author[2]{Ziyang Zhou}
\author[1$^\dag$]{Chi-Man Pun\thanks{$^\dag$ Corresponding author.}}

\affil[1]{University of Macau}
\affil[2]{Huizhou University}
\affil[3]{Shenzhen Institute of Advanced Technology, Chinese Academy of Sciences}

\maketitle

\begin{abstract}
Underwater image enhancement (UIE) plays a crucial role in various marine applications, but it remains challenging due to the complex underwater environment. Current learning-based approaches frequently lack explicit incorporation of prior knowledge about the physical processes involved in underwater image formation, resulting in limited optimization despite their impressive enhancement results. This paper proposes a novel deep unfolding network (DUN) for UIE that integrates color priors and inter-stage feature transformation to improve enhancement performance. The proposed DUN model combines the iterative optimization and reliability of model-based methods with the flexibility and representational power of deep learning, offering a more explainable and stable solution compared to existing learning-based UIE approaches. The proposed model consists of three key components: a Color Prior Guidance Block (CPGB) that establishes a mapping between color channels of degraded and original images, a Nonlinear Activation Gradient Descent Module (NAGDM) that simulates the underwater image degradation process, and an Inter Stage Feature Transformer (ISF-Former) that facilitates feature exchange between different network stages. By explicitly incorporating color priors and modeling the physical characteristics of underwater image formation, the proposed DUN model achieves more accurate and reliable enhancement results. Extensive experiments on multiple underwater image datasets demonstrate the superiority of the proposed model over state-of-the-art methods in both quantitative and qualitative evaluations. The proposed DUN-based approach offers a promising solution for UIE, enabling more accurate and reliable scientific analysis in marine research. The code is available at \href{https://github.com/CXH-Research/UIE-UnFold}{https://github.com/CXH-Research/UIE-UnFold}.
\end{abstract}


\section{Introduction}
The ocean is a vast and mysterious realm that covers more than 70\% of the Earth's surface. Underwater images provide scientists and researchers with valuable visual data, such as marine biology, underwater archaeology and underwater robotics~\cite{8706541}. However, acquiring clear and informative underwater images is full of challenges. The underwater environment poses unique obstacles that can significantly degrade the quality of the captured images. Light scattering, caused by the water itself and suspended particles, can result in images that appear hazy, blurry, and lacking in contrast. Additionally, the selective absorption of different wavelengths of light by water leads to color distortion, with red light being absorbed more quickly than blue and green light, resulting in images with limited color information. 

\begin{figure}[t]
    \begin{minipage}[b]{1.0\linewidth}
        \begin{minipage}[b]{.32\linewidth}
            \centering
            \centerline{\includegraphics[width=\linewidth]{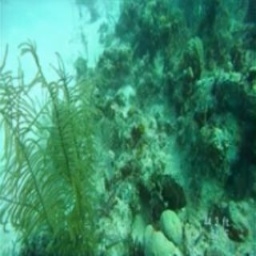}}
            \centerline{(a) Input}\medskip
        \end{minipage}
        \hfill
        \begin{minipage}[b]{.32\linewidth}
            \centering
            \centerline{\includegraphics[width=\linewidth]{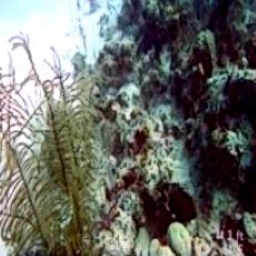}}
            \centerline{(b) HLRP}\medskip
        \end{minipage}
        \hfill
        \begin{minipage}[b]{0.32\linewidth}
            \centering
            \centerline{\includegraphics[width=\linewidth]{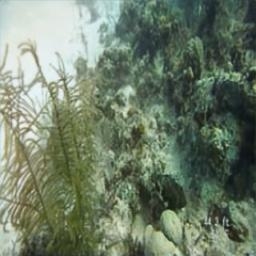}}
            \centerline{(c) CLUIE-Net}\medskip
        \end{minipage}
    \end{minipage}
    \begin{minipage}[b]{1.0\linewidth}
        \begin{minipage}[b]{.32\linewidth}
            \centering
            \centerline{\includegraphics[width=\linewidth]{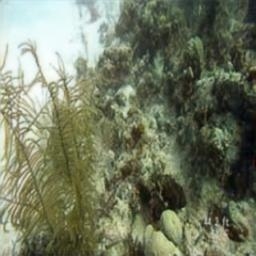}}
            \centerline{(d) PUGAN}\medskip
        \end{minipage}
        \hfill
        \begin{minipage}[b]{.32\linewidth}
            \centering
            \centerline{\includegraphics[width=\linewidth]{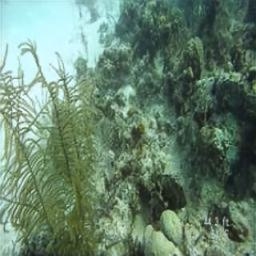}}
            \centerline{(e) Ours}\medskip
        \end{minipage}
        \hfill
        \begin{minipage}[b]{0.32\linewidth}
            \centering
            \centerline{\includegraphics[width=\linewidth]{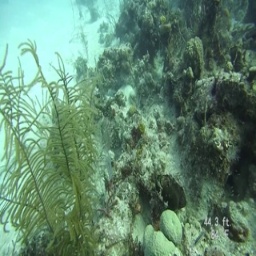}}
            \centerline{(f) Target}\medskip
        \end{minipage}
    \end{minipage}
    \caption{The underwater image enhancement results of our proposed model compared to several state-of-the-art methods on a sample underwater image. The input image (a) suffers from color distortion, low contrast, and haziness, which are typical issues in underwater imaging. Traditional methods like HLRP (b) struggle to effectively remove the haze and restore natural colors. Learning-based approaches such as CLUIE-Net (c) and PUGAN (d) show improved enhancement results but still exhibit some color deviations and artifacts. In contrast, our proposed model (e) successfully removes the haze, enhances the contrast, and restores vivid and natural colors, producing a visually pleasing result that closely resembles the target image (f). } 
    \label{fig:teaser}
\end{figure}

To mitigate these effects, underwater image enhancement (UIE) has emerged as a crucial tool. UIE aims to improve the visual quality and information content of underwater images, enabling more accurate and reliable scientific analysis. Traditional UIE methods, including model-based and model-free approaches~\cite{10.1007/s10462-021-10025-z}, have made significant contributions. However, these methods often require manual parameter tuning and may not generalize well to new underwater scenes or imaging conditions, as shown in Fig.~\ref{fig:teaser} (b). In recent years, learning-based UIE methods have gained significant attention as a promising solution to address the limitations of traditional UIE approaches. By training on large datasets, learning-based UIE methods can effectively restore visual quality across a wide range of underwater environments and imaging conditions, as shown in Fig.~\ref{fig:teaser} (c) and (d). 

While learning-based UIE methods have demonstrated impressive results in visual quality, they also have some limitations. As those based on convolutional neural networks (CNNs)~\cite{LI2020107038} or generative adversarial networks (GANs)~\cite{8460552}, typically learn the mapping between degraded and enhanced images directly from data, without explicitly incorporating prior knowledge about the physical processes involved in underwater image formation. Although data-driven learning methods demonstrate powerful pattern recognition and impressive enhancement results, they often lack the adaptability for iterative optimization.

Recently, the development of deep unfolding networks (DUN)~\cite{9157092}, which integrate model-based priors with learning-based flexibility, offers a promising direction for improving the optimization capabilities and reliability of UIE methods. DUNs bridge the gap between traditional model-based approaches and learning-based methods by unfolding iterative optimization algorithms into learnable deep neural networks. This allows DUNs to incorporate domain knowledge and physical priors while leveraging the representational power of deep learning. Inspired by the potential of DUNs, we propose a novel DUN-based model for UIE that introduces color priors of underwater images and integrates features between iteration stages to minimize feature loss. By explicitly modeling the physical characteristics of underwater image formation and enabling effective information flow within the network, our proposed model demonstrates strong performance and improved optimization capabilities compared to existing learning-based UIE methods. Overall, our contributions can be summarized as follows:
\begin{enumerate}
    \item We propose the Color Prior Guidance Block (CPGB), which implicitly establishes a mapping relationship between the color channels of the degraded and original images and introduces color priors to guide the image toward the color detail of the original image during the iteration process.
    \item We present the Nonlinear Activation Gradient Descent Module (NAGDM), employing a gradient estimation strategy to simulate the process involving multiple unknown parameters during underwater image degradation.
    \item We introduce the Inter Stage Feature Transformer (ISF-Former), utilizing the Transformer to extract image features and integrate them into the subsequent stage to facilitate the exchange and transformation of features between different stages of the network.
    \item Experimental results demonstrate that our model achieves excellent enhancement effects on images captured in various underwater conditions, showing superior performance over other state-of-the-art methods, thereby proving the effectiveness and superiority of our model.
\end{enumerate}

\section{Related Work}
\subsection{Traditional UIE Methods}
Traditional UIE methods are commonly classified into two primary categories: model-based methods and model-free methods.
\subsubsection{Model-based Methods}
Model-based underwater image enhancement methods rely on the principles of underwater image formation and degradation. These methods incorporate both physical models and image priors to estimate and compensate for the effects of light attenuation, scattering, and color distortion.
For instance, WCID~\cite{6104148} adopts the dark channel prior (DCP)~\cite{5206515} by introducing wavelength-dependent compensation. Similarly, UDCP~\cite{6755982} modifies the dark channel calculation to suit underwater scenes better. UNTV~\cite{9548907} incorporates the normalized total variation regularization, sparse prior knowledge of the blur kernel, and a red channel prior (RCP) based transmission map estimation into a unified variational framework to simultaneously remove haze and blur from underwater images. UIE-Blur~\cite{7351749} and IBLA~\cite{7840002} enhance underwater images by estimating depth from blurriness and using it with the image formation model to compensate for depth-dependent light attenuation and scattering. Min\_info\_loss~\cite{7574330} improves the quality of underwater images by removing the effects of haze while minimizing information loss and leveraging the prior knowledge of the image's histogram distribution. Model-based methods offer a principled approach, but their performance can be limited in dynamic underwater environments due to the simplified assumptions and the difficulty in accurately estimating model parameters.

\begin{figure*}[ht]
    \begin{minipage}[b]{1.0\linewidth}
        \begin{minipage}[b]{1.0\linewidth}
            \centering
            \centerline{\includegraphics[width=\linewidth]{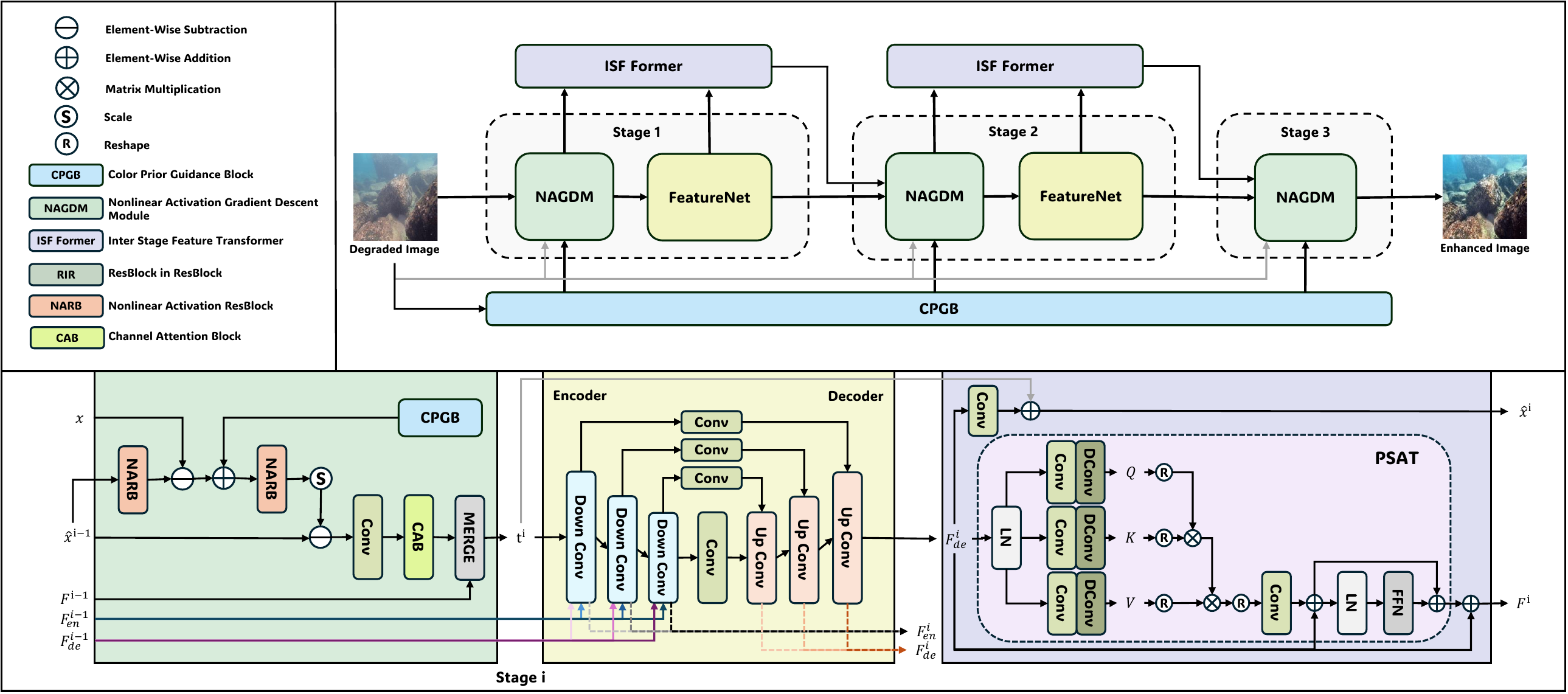}}
        \end{minipage}
    \end{minipage}
    \caption{The overall framework of the proposed deep unfolding network (DUN), namely UIE-UnFold, for underwater image enhancement (UIE). The framework consists of three main stages, each containing key components such as the Color Prior Guidance Block (CPGB), Nonlinear Activation Gradient Descent Module (NAGDM), FeatureNet, and Inter Stage Feature Transformer (ISF-Former).
    } 
    \label{fig:model}
\end{figure*}
\subsubsection{Model-free Methods}
Model-free methods aim to improve the visual quality of underwater images by directly manipulating the pixel values without relying on explicit physical or mathematical models. These methods employ various image processing techniques to adjust the color, contrast, brightness, and other visual attributes. For example, some approaches based on fusion strategy~\cite{6247661,10196309} leverage the complementary information from multiple input images (\eg color-corrected and contrast-enhanced images) and enhance the visual quality of underwater images through a certain fusion mechanism. Retinex-based methods like~\cite{7025927,ZHUANG2021104171} share the common approach of decomposing an image into illumination and reflectance components, then adjusting the illumination component to compensate for underwater light attenuation and enhancing the reflectance component to improve color and contrast.

\subsection{Learning-based UIE Methods}
Learning-based methods are also known as data-driven methods. By training on diverse pairs of degraded and clean images, these methods can capture the intricate patterns and relationships that characterize underwater image degradation, without the need for explicit physical modeling. 

Learning-based UIE methods have gained significant attention in recent years, with two prominent classes being CNN-based and GAN-based methods. CNNs-based methods utilize convolutional neural networks to learn the mapping between degraded and clean underwater images. For example, UWCNN~\cite{LI2020107038} is an end-to-end CNN-based model that enhances underwater images guided by underwater scene priors. UColor~\cite{9426457} introduces a multi-color space encoder network that leverages features from different color spaces, and the medium transmission-guided decoder network which emphasizes attention on degraded regions. CLUIE-Net~\cite{9965419} introduces a comparative learning framework that utilizes multi-reference learning strategy to capture a broader range of enhancement possibilities and adapt to various image contexts more effectively. GAN-based methods employ an adversarial training process where the generator network learns to create increasingly realistic and visually appealing enhanced underwater images. For instance, Fabbri \etal introduce UGAN~\cite{8460552} to translate degraded underwater images to clean ones and leverages CycleGAN to generate synthetic underwater image data for training purposes. PUGAN~\cite{10155564} introduces a dual-discriminator GAN architecture that incorporates a degradation parameter estimation module to guide the image enhancement process. 

\subsection{Deep Unfolding Networks}
Deep unfolding, proposed by Hershey \etal~\cite{hershey2014deep}, is a technique for converting conventional iterative algorithms into deep neural network architectures. This approach allows the integration of model-based knowledge into deep learning architectures. The unfolded network can be trained end-to-end using standard deep learning techniques, such as backpropagation and stochastic gradient descent. During training, the network learns to optimize the parameters of the unfolded algorithm, effectively adapting it to the specific task.

Deep unfolding has been successfully applied to a wide range of low-level vision tasks. For instance, Dong \etal~\cite{8481558} present a denoising-based image restoration algorithm that unfolds the traditional iterative optimization process into a deep neural network. Zhang \etal~\cite{9157092} introduce a deep unfolding super-resolution network for image super-resolution by unfolding the iterative optimization process. Zhang \etal~\cite{8578294} propose the ISTA-Net for image compressive sensing. It is designed by unfolding the iterative shrinkage-thresholding algorithm into a deep neural network architecture. 

Unfolding networks have great potential in UIE by integrating domain knowledge, offering robustness and generalization capabilities.

\begin{algorithm}
\caption{Model Forward Process}
\begin{algorithmic}[1]
\STATE \textbf{Input:} $img\in\mathbb{R}^{3 \times H \times W}$
\STATE \textbf{Output:} $[stage3_{img}, stage2_{img}, stage1_{img}]$
\STATE // Prior guided
\STATE Get prior from CPGB: $prior \gets \text{CPGM}(img)$
% \STATE 2: Feature extraction: $x1 \gets \text{res\_ch}(\text{concat}(tone\_x, tone\_x, tone\_x))$
% \STATE 2: Get prior from feature extraction: $prior_1 \gets \text{Conv}(\text{Concat}(tone_x, tone_x, tone_x))$
\FOR{$i \gets 1$ to $3$}
\STATE // Stage $i$
\STATE Apply NARB and integrate: $temp_i \gets \text{NARB}(stage(i-1)_{img}) - img + prior$
\STATE Scale after applying NARB and integrate: $x_i \gets stage(i-1)_{img} - scale \cdot \text{NARB}(prior_i)$
\IF{$i = 1$}
\STATE $enc_1, feat_1 \gets \text{Encoder}(x_1)$
\STATE $dec_1 \gets \text{Decoder}(enc_1, feat_1)$
\STATE Output: $feat_{12}, stage1_{img} \gets \text{ISF-Former}(dec_1, x_1)$
\ELSIF{$i = 2$}
\STATE Merge: $merged_{2} \gets \text{Merge}(x_2, feat_{12})$
\STATE $enc_2,feat_2 \gets \text{Encoder}(merged_{2}, enc_1, dec_1)$
\STATE $dec_2 \gets \text{Decoder}(feat_2, enc_2)$
\STATE Output: $feat_{23}, stage2_{img} \gets \text{ISF-Former}(dec_2, x_2)$
\ELSE
\STATE Merge: $merged_{3} \gets \text{Merge}(x_3, feat_{23})$
\STATE Output: $stage3_{img} \gets \text{Conv}(merged_{3}) + img$
\ENDIF
\ENDFOR
\STATE \textbf{return} $[stage3_{img}, stage2_{img}, stage1_{img}]$
\end{algorithmic}
\end{algorithm}

\section{Methodology}
The overall framework of our proposed model is illustrated in Fig.~\ref{fig:model}. The degraded input image is initially processed by the Color Prior Guidance Block (CPGB), which establishes a mapping between the color channels of the degraded and original images, providing color priors to guide the enhancement process. Each stage incorporates a Nonlinear Activation Gradient Descent Module (NAGDM), which simulates the underwater image degradation process using a gradient estimation strategy. 

This approach enhances the iterative optimization capabilities of the model while enabling end-to-end training. The Inter Stage Feature Transformer (ISF-Former) facilitates feature exchange and transformation between different stages of the network, allowing for more effective information flow and improving the overall enhancement performance. FeatureNet enables stage-specific feature extraction and transformation, supporting the iterative enhancement process and interaction between the NAGDM and ISFFormer modules. In the following subsections, we will introduce these modules in detail. 

\begin{figure}[ht]
\centering
    \begin{minipage}[b]{1.0\linewidth}
    \centering
        \begin{minipage}[b]{1.0\linewidth}
            \centering
            \centerline{\includegraphics[width=\linewidth]{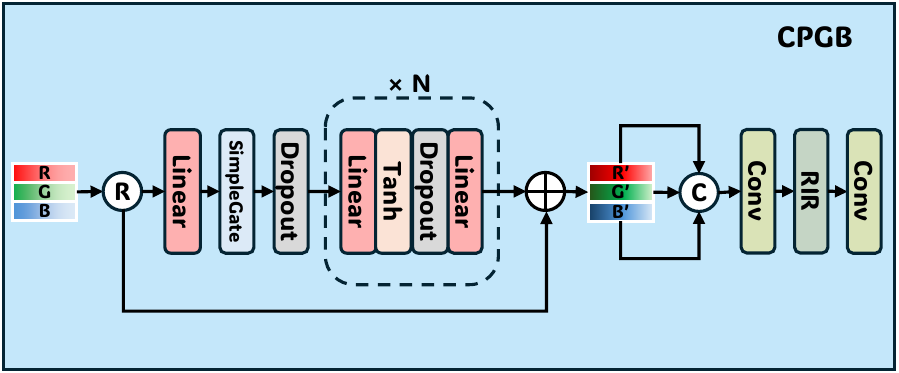}}
        \end{minipage}
    \end{minipage}
    \caption{The architecture of the Color Prior Guidance Block (CPGB). 
    % CPGB introduces color prior information to guide the image towards the color details of the original image during the iteration process by establishing a mapping relationship between the color channels of the degraded and original images.
    } 
    \label{fig:CPGB}
\end{figure}
\subsection{Color Prior Guidance Block (CPGB)}
The architecture of Color Prior Guidance Block is presented in Fig.~\ref{fig:CPGB}. CPGB is designed to implicitly establish a mapping relationship between the color channels of the degraded and original images, introducing color priors to guide the image toward the color detail of the original image during the iteration process.

Inspired by Implicit Neural Representations (INRs)~\cite{sitzmann2020implicit}, we utilize a a multilayer perceptron module to implicitly define a continuous 3-channel color transformation, which aims to establish a mapping between the input and target images:
\begin{equation}
    \Phi_{\theta}:x\in\mathbb{R}^{3 \times H \times W} \longmapsto y\in\mathbb{R}^{3 \times H \times W},
\end{equation}
assume we have a degraded underwater image $x$ with channel dimensional space being $(r, g, b)$, and our target image $y$ with channel dimensional space being $(r', g', b')$. During the training process, the model captures the mapping $\Phi_{\theta}$ by optimizing the distance between the two dimensional spaces.

\begin{figure}[ht]
\centering
    \begin{minipage}[b]{1.0\linewidth}
    \centering
        \begin{minipage}[b]{1.0\linewidth}
            \centering
            \centerline{\includegraphics[width=\linewidth]{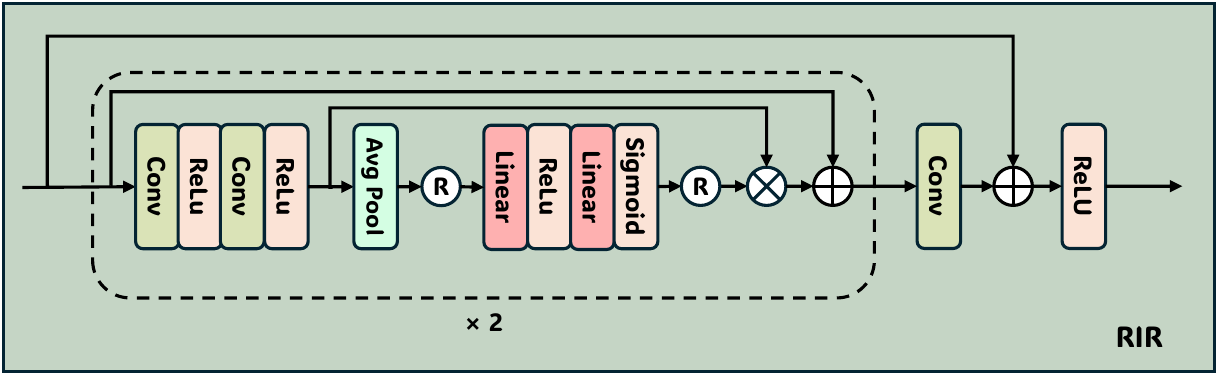}}
        \end{minipage}
    \end{minipage}
    \caption{The architecture of the ResBlock in ResBlock (RIR). 
    % RIR is a sub-module of CPGB that extracts color-channel features through a nested residual structure.
    } 
    \label{fig:RIR}
\end{figure}

Additionally, we adopt the ResBlock in ResBlock (RIR) in CPGB. The architecture of RIR is shown in Fig.~\ref{fig:RIR}. The RIR module consists of multiple stacked Residual Blocks (RBs), which can capture color features at different scales. The shallower RBs extract local color patterns, while the deeper RBs capture global color information. By incorporating these multi-scale color features, RIR enables CPGM to obtain a more comprehensive color representation. With the learned mapping $\Phi_{\theta}$, the final output is calculated as:
\begin{equation}
    \operatorname{CPGB}= \operatorname{Conv}(\operatorname{RIR}(\operatorname{Conv}(\Phi_{\theta}(x)+x))).
\end{equation}
\subsection{Nonlinear Activation Gradient Descent Module (NAGDM)}
Underwater images suffer from color distortion, low contrast, and haze-like effects due to the unique properties of underwater environments. The underwater image degradation model can be represented as follows:
\begin{equation}
\label{eq:degrad}
    y=\mathrm D x+\mathrm n,
\end{equation}
\begin{figure}[ht]
\centering
    \begin{minipage}[b]{1.0\linewidth}
    \centering
        \begin{minipage}[b]{1.0\linewidth}
            \centering
            \centerline{\includegraphics[width=\linewidth]{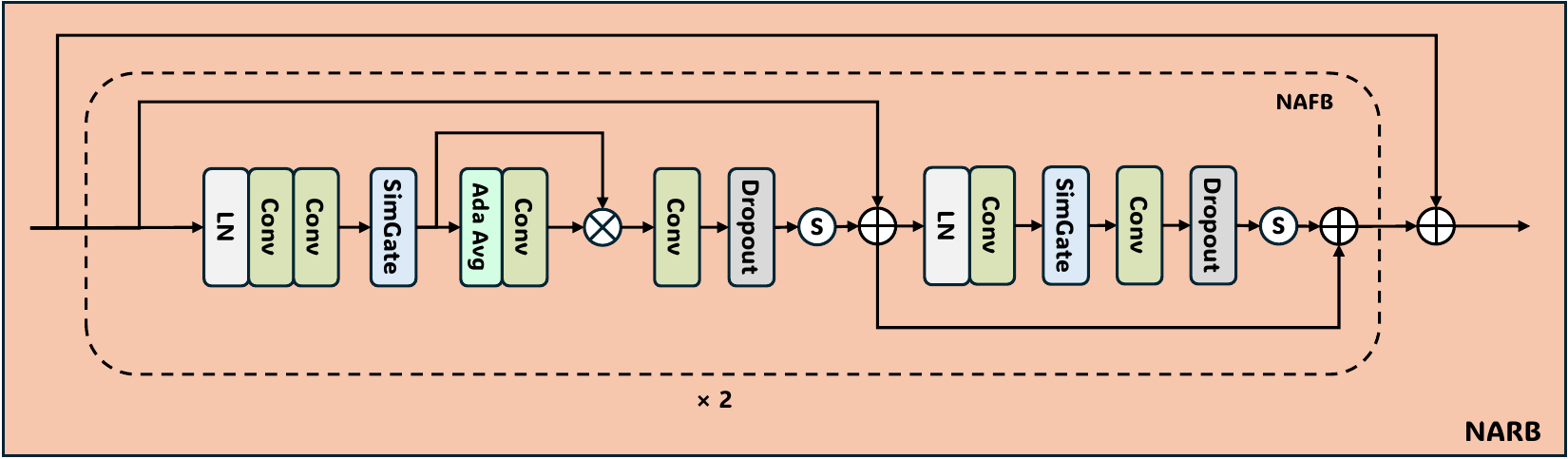}}
        \end{minipage}
    \end{minipage}
    \caption{The architecture of the Nonlinear Activation ResBlock (NARB). 
    % NARB is the basic component of the Nonlinear Activation Gradient Descent Module (NAGDM), used to simulate the process involving multiple unknown parameters during underwater image degradation.
    } 
    \label{fig:NARB}
\end{figure}
where $y$ is the observed degraded image, $x$ is the original clean image we aim to recover, $\mathrm D$ is the degradation matrix that models the degradation process, and $\mathrm n$ is the additive noise term. Traditional model-based methods usually regard UIE as a Bayesian problem, solving Eq.~\eqref{eq:degrad} from Bayesian maximum a posteriori perspective:
\begin{equation}
\label{eq:map}
    \hat{x}=\underset{x}{\mathrm{argmax}}\log P(x|y)=\underset{x}{\mathrm{argmax}}\log P(y|x)+\mathrm{log}P(x),
\end{equation}
where $\mathrm{log}P(y|x)$ represents the data fidelity and $\mathrm{log}P(x)$ represents regularization terms. Typically, the data fidelity is defined by the $l_2$ norm. Therefore, Eq.~\eqref{eq:map} can be formulated as:
\begin{equation}
\label{eq:5}
    \hat{x}=\underset{x}{\operatorname*{argmin}}\frac{1}{2}||y-\mathrm{D}x||_2^2+\lambda J(x),
\end{equation}
where $\lambda$ serves as a hyper-parameter to adjust the impact of the regularization term $J(x)$. Based on the idea of transforming the optimization problem in Eq.~\eqref{eq:5} into an iterative convergence problem, the Proximal Gradient Descent (PGD) algorithm approximates the solution through the following iterative process~\cite{9878586}:
\begin{equation}
    \hat{x}^i=\underset{\mathbf{x}}{\operatorname*{argmin}}\frac12||x-(\hat{x}^{i-1}-\omega\nabla g(\hat{x}^{i-1}))||_2^2+\lambda J(x),
\end{equation}
where the first component corresponds to the gradient descent step, and the second component corresponds to the proximal mapping step. Consequently, the problem can be split into two smaller subproblems:
\begin{equation}
\begin{aligned}
    t^i&=\hat{x}^{i-1}-\omega\mathrm{D}^\top(\mathrm{D}\hat{x}^{i-1}-y),\\
    \hat{x}^i&=\operatorname{prox}_{\lambda,J}(t^i).
\end{aligned}
\end{equation}

Underwater image degradation matrices are complex and inconsistent across different underwater environments. To address this challenge, NAGDM employs a gradient estimation strategy to simulate the process involving multiple unknown parameters during underwater image degradation. The framework of NAGDM is depicted in the green block in Fig.~\ref{fig:model}. 

Inspired by~\cite{10.1007/978-3-031-20071-7_2,chen2,chen3,chen4}, we propose a Nonlinear Activation ResBlock to learn the complex and varied degradation matrices in UIE without the need for manual definition. The NARB module is designed to simulate the degradation matrix. It enables the learning of flexible nonlinear transformations, incorporates an adaptive gating mechanism, facilitates end-to-end training, and benefits from residual learning. The architecture of NARB is presented in Fig.~\ref{fig:NARB}.
We utilize two NAGDM to present the degradation matrices $\mathrm{D}$ and $\mathrm{D}^\top$ as $\mathcal{N}_{\mathrm{D}}^{i}$ and $\mathcal{N}_{\mathbf{D}^{\top}}^{i}$, respectively. The iteration process can be formulated as:
\begin{equation}
    t^i=\hat{x}^{i-1}-\omega\mathcal{N}_{\mathbf{D}^{\top}}^{i}(\mathcal{N}_{\mathrm{D}}^{i}(\hat{x}^{i-1})-y).
\end{equation}

\subsection{Inter Stage Feature Transformer (ISF-Former)}
Inspired by~\cite{Zamir_Arora_Khan_Hayat_Khan_Yang_Shao_2021,chen1,chen6,chen7,chen8}, we introduce the Inter Stage Feature Transformer to facilitate feature exchange and transformation between different stages of the deep unfolding network. The architecture of ISF-Former is presented in the purple block in Fig.~\ref{fig:model}.

Transformers are known for their ability to capture long-range dependencies and global context. By employing a Transformer-based module, the ISF-Former can extract and integrate features that encode global information about the image, leading to more coherent and contextually aware enhancements.

ISF-Former is mainly based on Pixel Self-Attention Transformer (PSAT). Taking $F_{de}^i \in\mathbb{R}^{C\times{H}\times{W}}$ from FeatureNet, we first calculate $Q, K, V$ as follows:
\begin{equation}
     Q,K,V=\operatorname{Conv}(\operatorname{DWConv}(F_{de}^i)),   
\end{equation}
where $Q, K, V\in\mathbb{R}^{C\times{H}\times{W}}$ are query, key, and value respectively, and then the pixel self-attention~\cite{chen5,chen9,chen10,chen11} can be formulated as follows:
\begin{equation}
    \operatorname{Attn}=V\cdot \operatorname{Softmax}(\frac{QK^T}{\sqrt{S}}),
\end{equation}
where $S$ presents the spatial size of the input feature, and then PSAT can be presented as follows:
\begin{equation}
\begin{aligned}
    F^{'}_{temp}&=\operatorname{Attn}(\operatorname{LN}(F_{de}^i))+F_{de}^i,\\
    F^{'}_{out}&=\operatorname{FFN}(\operatorname{LN}(F^{'}_{temp}))+F^{'}_{temp},
\end{aligned}
\end{equation}
where $\operatorname{LN}$ represents LayerNorm and $\operatorname{FFN}$ means Feed Forward Network. The final output feature is calculated as:
\begin{equation}
    F^i=F^i_{de}+F^{'}_{out},
\end{equation}
where $F^i\in\mathbb{R}^{C\times{H}\times{W}}$.
The integration of features from one stage to the next allows for progressive refinement of the enhanced image. Each stage can build upon the features extracted by the previous stage, enabling the network to generate high-quality enhancements iteratively.

\subsection{FeatureNet}
Similar to~\cite{9878586}, we adopt a U-shaped encoder-decoder network as FeatureNet, as represented in Fig.~\ref{fig:model}. The FeatureNet architecture is designed to extract meaningful features from input images and reconstruct them back to the original resolution. 

The encoder progressively reduces the spatial dimensions of the input while increasing the depth, allowing the network to capture high-level features. This is achieved through a series of convolutional layers and downsampling operations. On the other hand, the decoder aims to reconstruct the encoded features back to the original spatial resolution. It employs transposed convolutional layers to gradually increase the spatial dimensions while reducing the depth.

To enhance the information flow and preserve spatial details, skip connections are incorporated into the FeatureNet architecture. These connections allow information from earlier layers in the encoder to be directly passed to the corresponding layers in the decoder. By leveraging skip connections, the network can effectively combine low-level and high-level features, leading to improved reconstruction quality.

\begin{figure*}[!ht]
    \begin{minipage}[b]{1.0\linewidth}
        \begin{minipage}[b]{0.105\linewidth}
            \centering
            \centerline{\includegraphics[width=\linewidth]{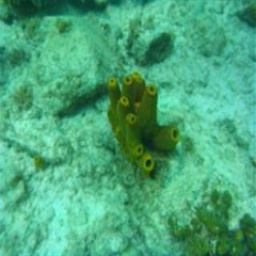}}
        \end{minipage}
        \hfill
        \begin{minipage}[b]{0.105\linewidth}
            \centering
            \centerline{\includegraphics[width=\linewidth]{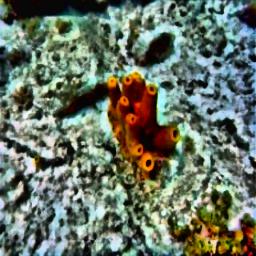}}
        \end{minipage}
        \hfill
        \begin{minipage}[b]{0.105\linewidth}
            \centering
            \centerline{\includegraphics[width=\linewidth]{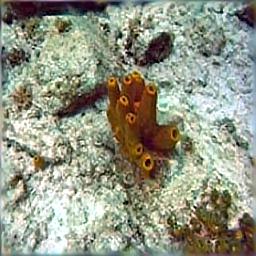}}
        \end{minipage}
        \hfill
        \begin{minipage}[b]{0.105\linewidth}
            \centering
            \centerline{\includegraphics[width=\linewidth]{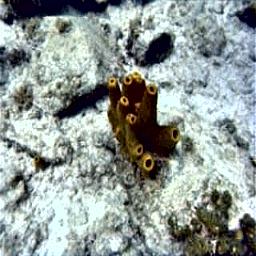}}
        \end{minipage}
        \hfill
        \begin{minipage}[b]{0.105\linewidth}
            \centering
            \centerline{\includegraphics[width=\linewidth]{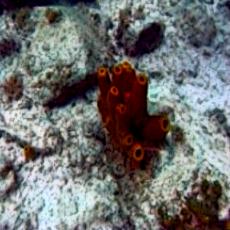}}
        \end{minipage}
        \hfill
        \begin{minipage}[b]{0.105\linewidth}
            \centering
            \centerline{\includegraphics[width=\linewidth]{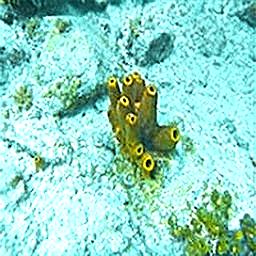}}
        \end{minipage}
        \hfill
        \begin{minipage}[b]{0.105\linewidth}
            \centering
            \centerline{\includegraphics[width=\linewidth]{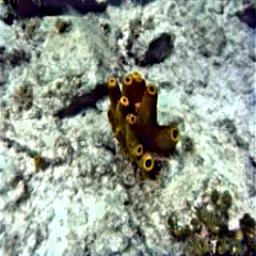}}
        \end{minipage}
        \hfill
        \begin{minipage}[b]{0.105\linewidth}
            \centering
            \centerline{\includegraphics[width=\linewidth]{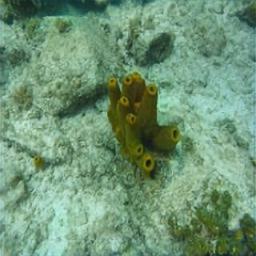}}
        \end{minipage}
        \hfill
        \begin{minipage}[b]{0.105\linewidth}
            \centering
            \centerline{\includegraphics[width=\linewidth]{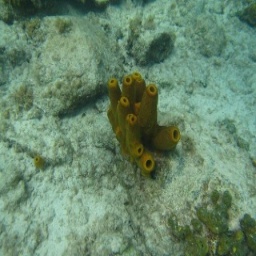}}
        \end{minipage}
    \end{minipage}

    \begin{minipage}[b]{1.0\linewidth}
        \begin{minipage}[b]{0.105\linewidth}
            \centering
            \centerline{\includegraphics[width=\linewidth]{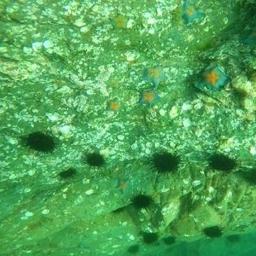}}
        \end{minipage}
        \hfill
        \begin{minipage}[b]{0.105\linewidth}
            \centering
            \centerline{\includegraphics[width=\linewidth]{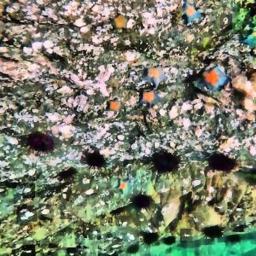}}
        \end{minipage}
        \hfill
        \begin{minipage}[b]{0.105\linewidth}
            \centering
            \centerline{\includegraphics[width=\linewidth]{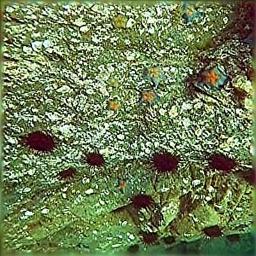}}
        \end{minipage}
        \hfill
        \begin{minipage}[b]{0.105\linewidth}
            \centering
            \centerline{\includegraphics[width=\linewidth]{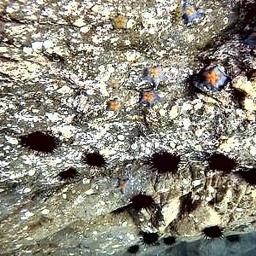}}
        \end{minipage}
        \hfill
        \begin{minipage}[b]{0.105\linewidth}
            \centering
            \centerline{\includegraphics[width=\linewidth]{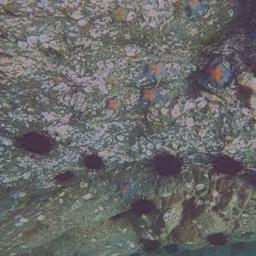}}
        \end{minipage}
        \hfill
        \begin{minipage}[b]{0.105\linewidth}
            \centering
            \centerline{\includegraphics[width=\linewidth]{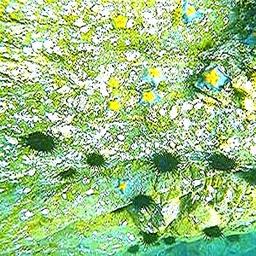}}
        \end{minipage}
        \hfill
        \begin{minipage}[b]{0.105\linewidth}
            \centering
            \centerline{\includegraphics[width=\linewidth]{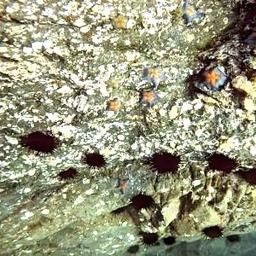}}
        \end{minipage}
        \hfill
        \begin{minipage}[b]{0.105\linewidth}
            \centering
            \centerline{\includegraphics[width=\linewidth]{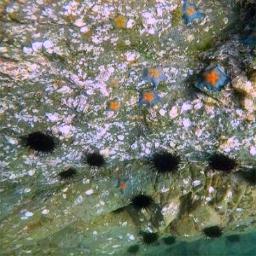}}
        \end{minipage}
        \hfill
        \begin{minipage}[b]{0.105\linewidth}
            \centering
            \centerline{\includegraphics[width=\linewidth]{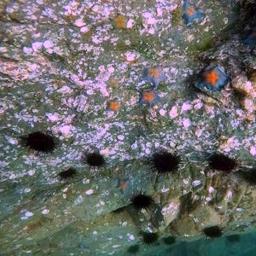}}
        \end{minipage}
    \end{minipage}

    \begin{minipage}[b]{1.0\linewidth}
        \begin{minipage}[b]{0.105\linewidth}
            \centering
            \centerline{\includegraphics[width=\linewidth]{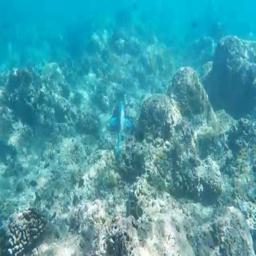}}
        \end{minipage}
        \hfill
        \begin{minipage}[b]{0.105\linewidth}
            \centering
            \centerline{\includegraphics[width=\linewidth]{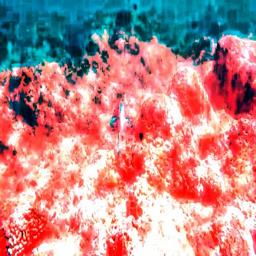}}
        \end{minipage}
        \hfill
        \begin{minipage}[b]{0.105\linewidth}
            \centering
            \centerline{\includegraphics[width=\linewidth]{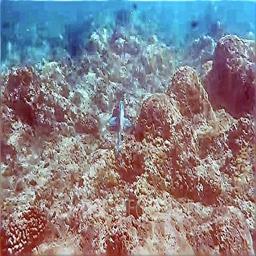}}
        \end{minipage}
        \hfill
        \begin{minipage}[b]{0.105\linewidth}
            \centering
            \centerline{\includegraphics[width=\linewidth]{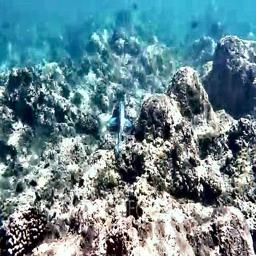}}
        \end{minipage}
        \hfill
        \begin{minipage}[b]{0.105\linewidth}
            \centering
            \centerline{\includegraphics[width=\linewidth]{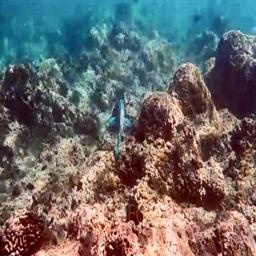}}
        \end{minipage}
        \hfill
        \begin{minipage}[b]{0.105\linewidth}
            \centering
            \centerline{\includegraphics[width=\linewidth]{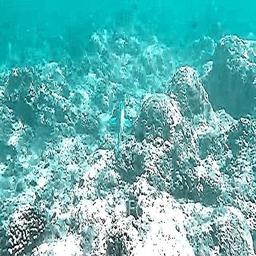}}
        \end{minipage}
        \hfill
        \begin{minipage}[b]{0.105\linewidth}
            \centering
            \centerline{\includegraphics[width=\linewidth]{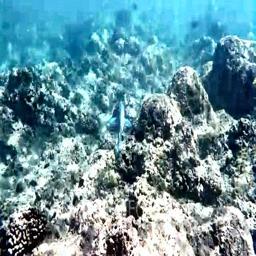}}
        \end{minipage}
        \hfill
        \begin{minipage}[b]{0.105\linewidth}
            \centering
            \centerline{\includegraphics[width=\linewidth]{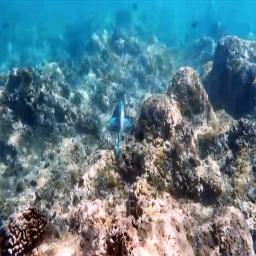}}
        \end{minipage}
        \hfill
        \begin{minipage}[b]{0.105\linewidth}
            \centering
            \centerline{\includegraphics[width=\linewidth]{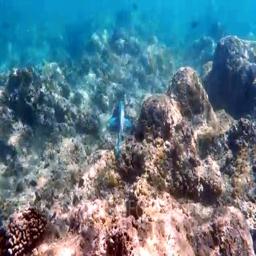}}
        \end{minipage}
    \end{minipage}

    \begin{minipage}[b]{1.0\linewidth}
        \begin{minipage}[b]{0.105\linewidth}
            \centering
            \centerline{\includegraphics[width=\linewidth]{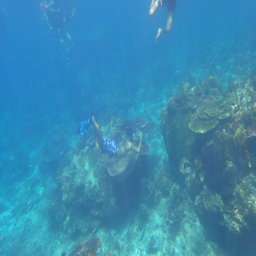}}
        \end{minipage}
        \hfill
        \begin{minipage}[b]{0.105\linewidth}
            \centering
            \centerline{\includegraphics[width=\linewidth]{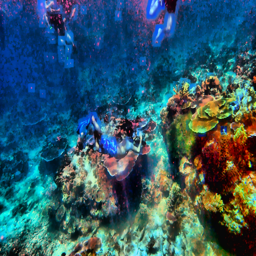}}
        \end{minipage}
        \hfill
        \begin{minipage}[b]{0.105\linewidth}
            \centering
            \centerline{\includegraphics[width=\linewidth]{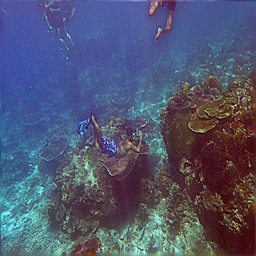}}
        \end{minipage}
        \hfill
        \begin{minipage}[b]{0.105\linewidth}
            \centering
            \centerline{\includegraphics[width=\linewidth]{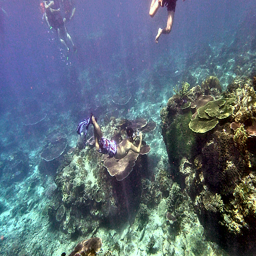}}
        \end{minipage}
        \hfill
        \begin{minipage}[b]{0.105\linewidth}
            \centering
            \centerline{\includegraphics[width=\linewidth]{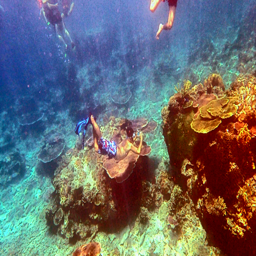}}
        \end{minipage}
        \hfill
        \begin{minipage}[b]{0.105\linewidth}
            \centering
            \centerline{\includegraphics[width=\linewidth]{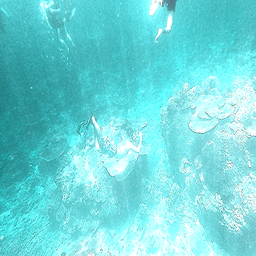}}
        \end{minipage}
        \hfill
        \begin{minipage}[b]{0.105\linewidth}
            \centering
            \centerline{\includegraphics[width=\linewidth]{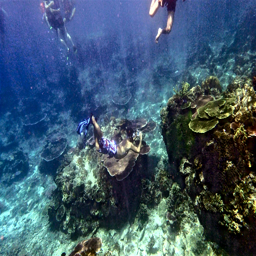}}
        \end{minipage}
        \hfill
        \begin{minipage}[b]{0.105\linewidth}
            \centering
            \centerline{\includegraphics[width=\linewidth]{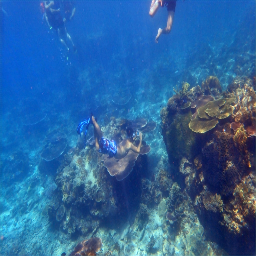}}
        \end{minipage}
        \hfill
        \begin{minipage}[b]{0.105\linewidth}
            \centering
            \centerline{\includegraphics[width=\linewidth]{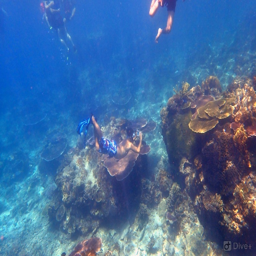}}
        \end{minipage}
    \end{minipage}
    
    \begin{minipage}[b]{1.0\linewidth}
        \begin{minipage}[b]{0.105\linewidth}
            \centering
            \centerline{\includegraphics[width=\linewidth]{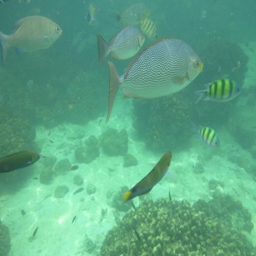}}
            \centerline{(a) Input}\medskip
        \end{minipage}
        \hfill
        \begin{minipage}[b]{0.105\linewidth}
            \centering
            \centerline{\includegraphics[width=\linewidth]{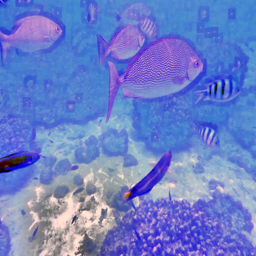}}
            \centerline{(b) Sea-thru}\medskip
        \end{minipage}
        \hfill
        \begin{minipage}[b]{0.105\linewidth}
            \centering
            \centerline{\includegraphics[width=\linewidth]{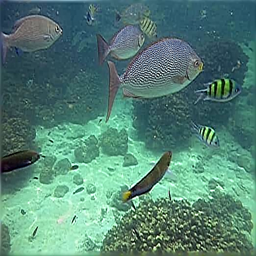}}
            \centerline{(c) UNTV}\medskip
        \end{minipage}
        \hfill
        \begin{minipage}[b]{0.105\linewidth}
            \centering
            \centerline{\includegraphics[width=\linewidth]{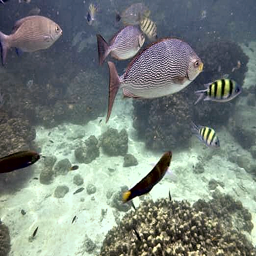}}
            \centerline{(d) MLLE}\medskip
        \end{minipage}
        \hfill
        \begin{minipage}[b]{0.105\linewidth}
            \centering
            \centerline{\includegraphics[width=\linewidth]{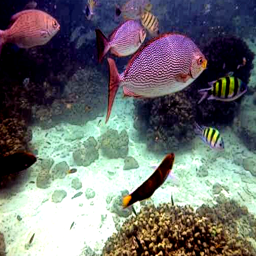}}
            \centerline{(e) HLRP}\medskip
        \end{minipage}
        \hfill
        \begin{minipage}[b]{0.105\linewidth}
            \centering
            \centerline{\includegraphics[width=\linewidth]{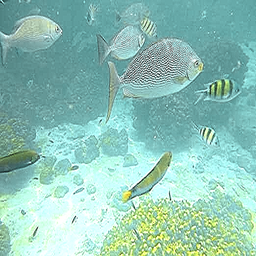}}
            \centerline{(f) ICSP}\medskip
        \end{minipage}
        \hfill
        \begin{minipage}[b]{0.105\linewidth}
            \centering
            \centerline{\includegraphics[width=\linewidth]{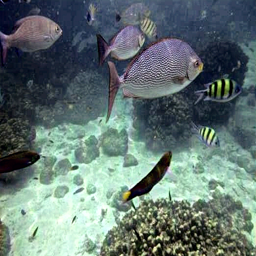}}
            \centerline{(g) WWPF}\medskip
        \end{minipage}
        \hfill
        \begin{minipage}[b]{0.105\linewidth}
            \centering
            \centerline{\includegraphics[width=\linewidth]{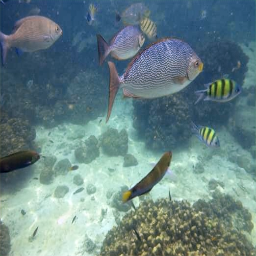}}
            \centerline{(h) Ours}\medskip
        \end{minipage}
        \hfill
        \begin{minipage}[b]{0.105\linewidth}
            \centering
            \centerline{\includegraphics[width=\linewidth]{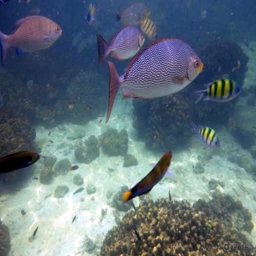}}
            \centerline{(i) Target}\medskip
        \end{minipage}
    \end{minipage}

    \caption{The qualitative comparison results between the proposed method and traditional UIE methods. 
    % It can be seen that the proposed method outperforms traditional methods in dehazing, color restoration, and contrast enhancement. Moreover, compared to traditional methods, our proposed method demonstrates strong adaptability to various underwater environments, exhibiting broad applicability.
    }
    \label{fig:compare_tradition}
    
\end{figure*}
\begin{figure*}[!ht]
    \begin{minipage}[b]{1.0\linewidth}
        \begin{minipage}[b]{0.105\linewidth}
            \centering
            \centerline{\includegraphics[width=\linewidth]{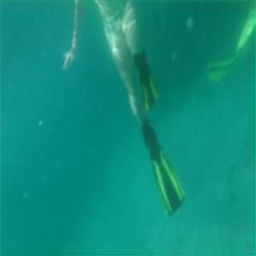}}
        \end{minipage}
        \hfill
        \begin{minipage}[b]{0.105\linewidth}
            \centering
            \centerline{\includegraphics[width=\linewidth]{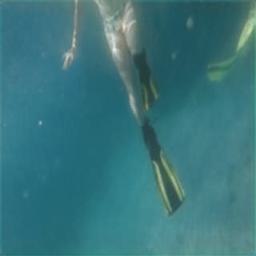}}
        \end{minipage}
        \hfill
        \begin{minipage}[b]{0.105\linewidth}
            \centering
            \centerline{\includegraphics[width=\linewidth]{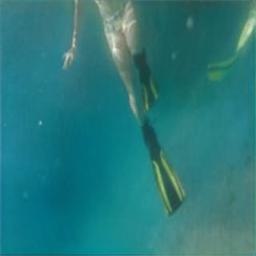}}
        \end{minipage}
        \hfill
        \begin{minipage}[b]{0.105\linewidth}
            \centering
            \centerline{\includegraphics[width=\linewidth]{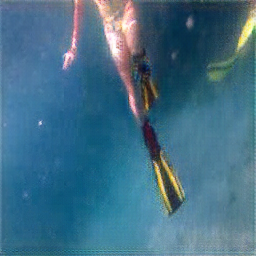}}
        \end{minipage}
        \hfill
        \begin{minipage}[b]{0.105\linewidth}
            \centering
            \centerline{\includegraphics[width=\linewidth]{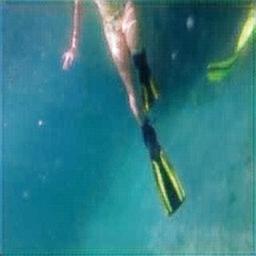}}
        \end{minipage}
        \hfill
        \begin{minipage}[b]{0.105\linewidth}
            \centering
            \centerline{\includegraphics[width=\linewidth]{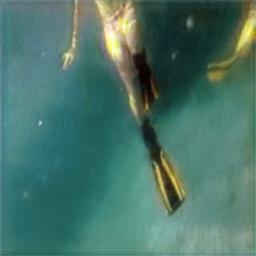}}
        \end{minipage}
        \hfill
        \begin{minipage}[b]{0.105\linewidth}
            \centering
            \centerline{\includegraphics[width=\linewidth]{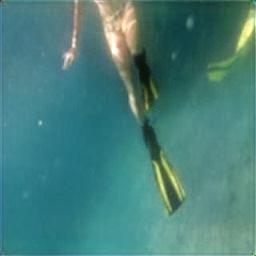}}
        \end{minipage}
        \hfill
        \begin{minipage}[b]{0.105\linewidth}
            \centering
            \centerline{\includegraphics[width=\linewidth]{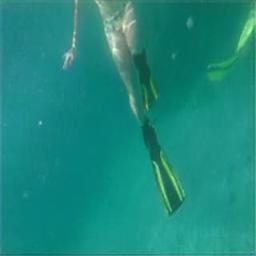}}
        \end{minipage}
        \hfill
        \begin{minipage}[b]{0.105\linewidth}
            \centering
            \centerline{\includegraphics[width=\linewidth]{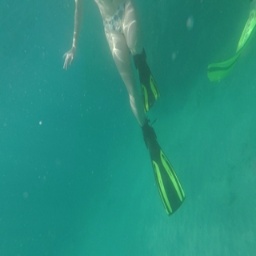}}
        \end{minipage}
    \end{minipage}

    \begin{minipage}[b]{1.0\linewidth}
        \begin{minipage}[b]{0.105\linewidth}
            \centering
            \centerline{\includegraphics[width=\linewidth]{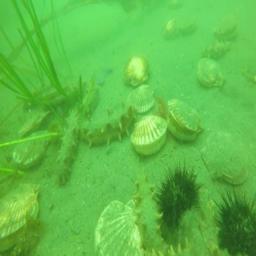}}
        \end{minipage}
        \hfill
        \begin{minipage}[b]{0.105\linewidth}
            \centering
            \centerline{\includegraphics[width=\linewidth]{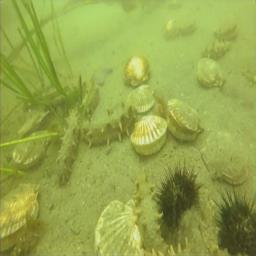}}
        \end{minipage}
        \hfill
        \begin{minipage}[b]{0.105\linewidth}
            \centering
            \centerline{\includegraphics[width=\linewidth]{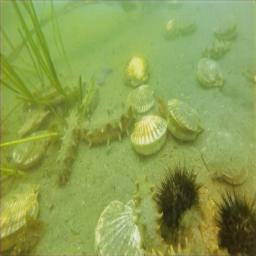}}
        \end{minipage}
        \hfill
        \begin{minipage}[b]{0.105\linewidth}
            \centering
            \centerline{\includegraphics[width=\linewidth]{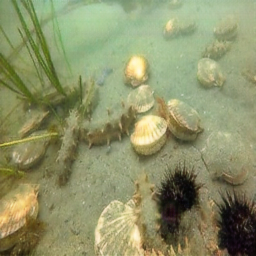}}
        \end{minipage}
        \hfill
        \begin{minipage}[b]{0.105\linewidth}
            \centering
            \centerline{\includegraphics[width=\linewidth]{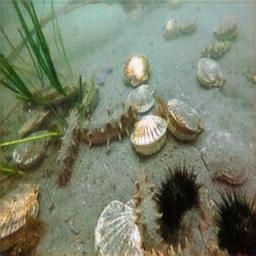}}
        \end{minipage}
        \hfill
        \begin{minipage}[b]{0.105\linewidth}
            \centering
            \centerline{\includegraphics[width=\linewidth]{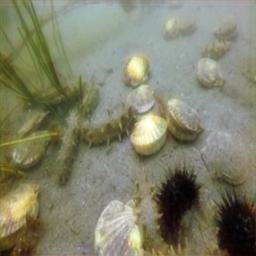}}
        \end{minipage}
        \hfill
        \begin{minipage}[b]{0.105\linewidth}
            \centering
            \centerline{\includegraphics[width=\linewidth]{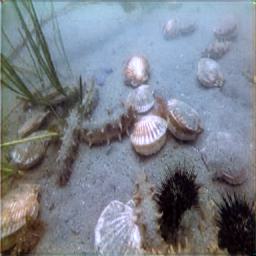}}
        \end{minipage}
        \hfill
        \begin{minipage}[b]{0.105\linewidth}
            \centering
            \centerline{\includegraphics[width=\linewidth]{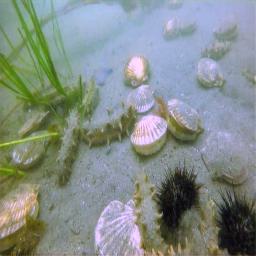}}
        \end{minipage}
        \hfill
        \begin{minipage}[b]{0.105\linewidth}
            \centering
            \centerline{\includegraphics[width=\linewidth]{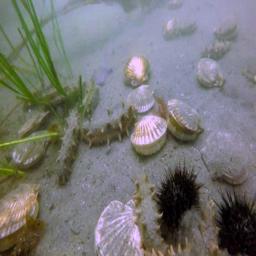}}
        \end{minipage}
    \end{minipage}

    \begin{minipage}[b]{1.0\linewidth}
        \begin{minipage}[b]{0.105\linewidth}
            \centering
            \centerline{\includegraphics[width=\linewidth]{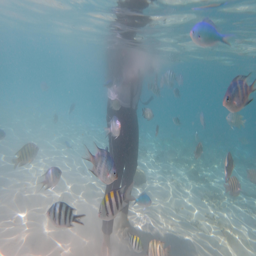}}
        \end{minipage}
        \hfill
        \begin{minipage}[b]{0.105\linewidth}
            \centering
            \centerline{\includegraphics[width=\linewidth]{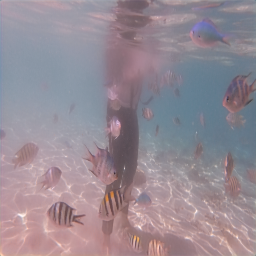}}
        \end{minipage}
        \hfill
        \begin{minipage}[b]{0.105\linewidth}
            \centering
            \centerline{\includegraphics[width=\linewidth]{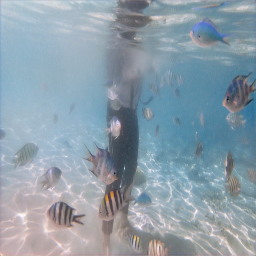}}
        \end{minipage}
        \hfill
        \begin{minipage}[b]{0.105\linewidth}
            \centering
            \centerline{\includegraphics[width=\linewidth]{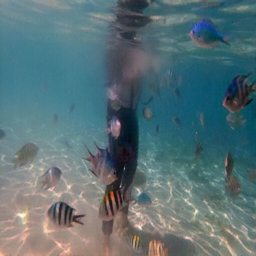}}
        \end{minipage}
        \hfill
        \begin{minipage}[b]{0.105\linewidth}
            \centering
            \centerline{\includegraphics[width=\linewidth]{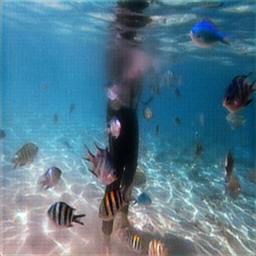}}
        \end{minipage}
        \hfill
        \begin{minipage}[b]{0.105\linewidth}
            \centering
            \centerline{\includegraphics[width=\linewidth]{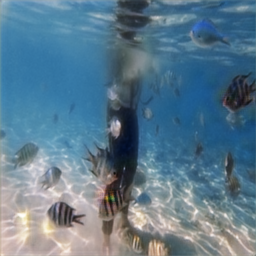}}
        \end{minipage}
        \hfill
        \begin{minipage}[b]{0.105\linewidth}
            \centering
            \centerline{\includegraphics[width=\linewidth]{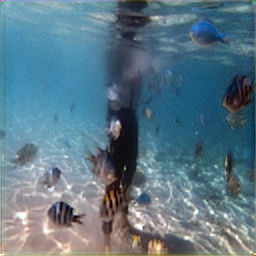}}
        \end{minipage}
        \hfill
        \begin{minipage}[b]{0.105\linewidth}
            \centering
            \centerline{\includegraphics[width=\linewidth]{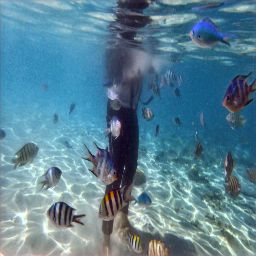}}
        \end{minipage}
        \hfill
        \begin{minipage}[b]{0.105\linewidth}
            \centering
            \centerline{\includegraphics[width=\linewidth]{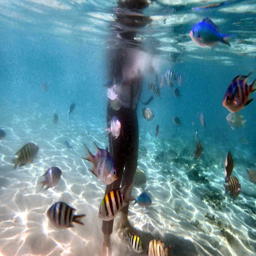}}
        \end{minipage}
    \end{minipage}

    \begin{minipage}[b]{1.0\linewidth}
        \begin{minipage}[b]{0.105\linewidth}
            \centering
            \centerline{\includegraphics[width=\linewidth]{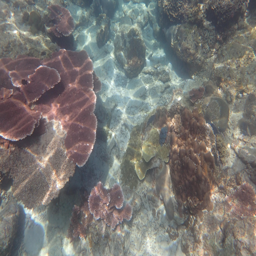}}
        \end{minipage}
        \hfill
        \begin{minipage}[b]{0.105\linewidth}
            \centering
            \centerline{\includegraphics[width=\linewidth]{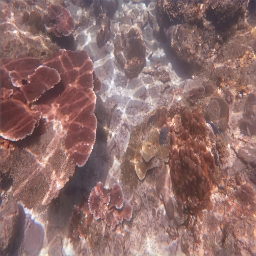}}
        \end{minipage}
        \hfill
        \begin{minipage}[b]{0.105\linewidth}
            \centering
            \centerline{\includegraphics[width=\linewidth]{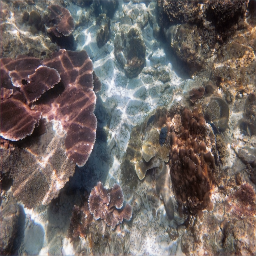}}
        \end{minipage}
        \hfill
        \begin{minipage}[b]{0.105\linewidth}
            \centering
            \centerline{\includegraphics[width=\linewidth]{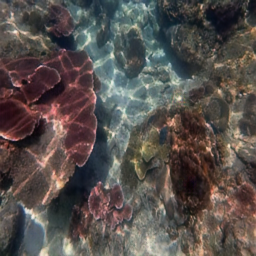}}
        \end{minipage}
        \hfill
        \begin{minipage}[b]{0.105\linewidth}
            \centering
            \centerline{\includegraphics[width=\linewidth]{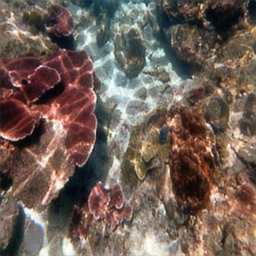}}
        \end{minipage}
        \hfill
        \begin{minipage}[b]{0.105\linewidth}
            \centering
            \centerline{\includegraphics[width=\linewidth]{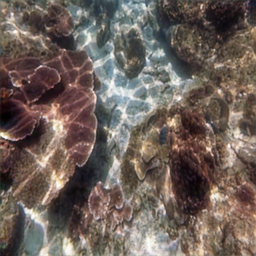}}
        \end{minipage}
        \hfill
        \begin{minipage}[b]{0.105\linewidth}
            \centering
            \centerline{\includegraphics[width=\linewidth]{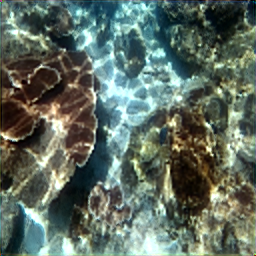}}
        \end{minipage}
        \hfill
        \begin{minipage}[b]{0.105\linewidth}
            \centering
            \centerline{\includegraphics[width=\linewidth]{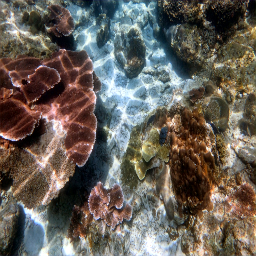}}
        \end{minipage}
        \hfill
        \begin{minipage}[b]{0.105\linewidth}
            \centering
            \centerline{\includegraphics[width=\linewidth]{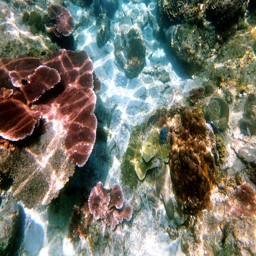}}
        \end{minipage}
    \end{minipage}

    \begin{minipage}[b]{1.0\linewidth}
        \begin{minipage}[b]{0.105\linewidth}
            \centering
            \centerline{\includegraphics[width=\linewidth]{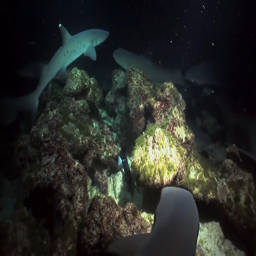}}
            \centerline{(a) Input}\medskip
        \end{minipage}
        \hfill
        \begin{minipage}[b]{0.105\linewidth}
            \centering
            \centerline{\includegraphics[width=\linewidth]{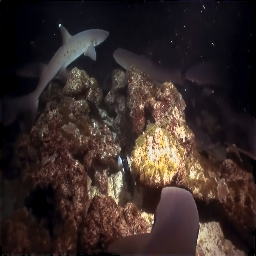}}
            \centerline{(b) UColor}\medskip
        \end{minipage}
        \hfill
        \begin{minipage}[b]{0.105\linewidth}
            \centering
            \centerline{\includegraphics[width=\linewidth]{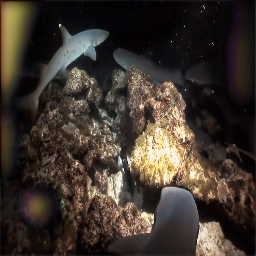}}
            \centerline{(c) CLUIE-Net}\medskip
        \end{minipage}
        \hfill
        \begin{minipage}[b]{0.105\linewidth}
            \centering
            \centerline{\includegraphics[width=\linewidth]{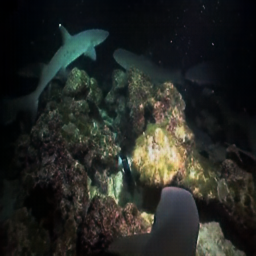}}
            \centerline{(d) TACL}\medskip
        \end{minipage}
        \hfill
        \begin{minipage}[b]{0.105\linewidth}
            \centering
            \centerline{\includegraphics[width=\linewidth]{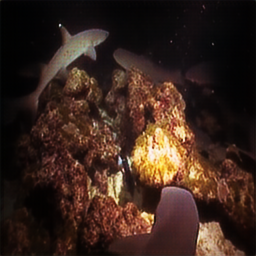}}
            \centerline{(e) UGAN}\medskip
        \end{minipage}
        \hfill
        \begin{minipage}[b]{0.105\linewidth}
            \centering
            \centerline{\includegraphics[width=\linewidth]{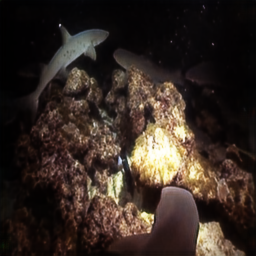}}
            \centerline{(f) PUGAN}\medskip
        \end{minipage}
        \hfill
        \begin{minipage}[b]{0.105\linewidth}
            \centering
            \centerline{\includegraphics[width=\linewidth]{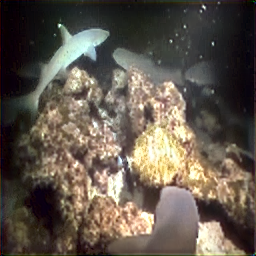}}
            \centerline{(g) U-Trans}\medskip
        \end{minipage}
        \hfill
        \begin{minipage}[b]{0.105\linewidth}
            \centering
            \centerline{\includegraphics[width=\linewidth]{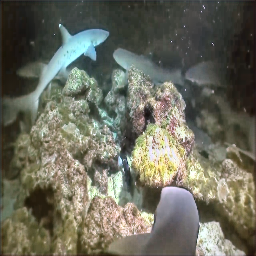}}
            \centerline{(h) Ours}\medskip
        \end{minipage}
        \hfill
        \begin{minipage}[b]{0.105\linewidth}
            \centering
            \centerline{\includegraphics[width=\linewidth]
            {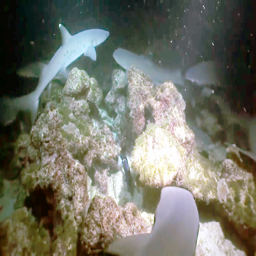}}
            \centerline{(i) Target}\medskip
        \end{minipage}
    \end{minipage}

    \caption{The qualitative comparison results between the proposed method and learning-based UIE methods. 
    % The proposed method not only surpasses other methods in texture detail restoration but also excels in color saturation and contrast enhancement.
    }
    \label{fig:compare_learningbased}
    
\end{figure*}
\begin{table*}[!ht]
    \centering
    \caption{Quantitative results of comparative experiments with state-of-the-art methods. {\color[HTML]{FE0000} Red} and \textbf{bold} font indicates the best results. The underlined text indicates the second-best results. }
    \begin{adjustbox}{max width=\textwidth}
    \begin{tabular}{c|cccc|cccc|cccc}
    \toprule
    \multirow{2}{*}{Method}&\multicolumn{4}{c|}{UIEB}&\multicolumn{4}{c|}{EUVP}&\multicolumn{4}{c}{LSUI}
    \\
    \cmidrule(l){2-13}
    & PSNR $\uparrow$ & SSIM $\uparrow$ & $\Delta$E $\downarrow$ & LPIPS $\downarrow$
    & PSNR $\uparrow$ & SSIM $\uparrow$ & $\Delta$E $\downarrow$ & LPIPS $\downarrow$
    & PSNR $\uparrow$ & SSIM $\uparrow$ & $\Delta$E $\downarrow$ & LPIPS $\downarrow$
    \\
    \midrule
    min\_info\_loss~\cite{7574330}  & 17.130 & 0.783 & 17.357 & 0.327 & 15.234 & 0.643 & 20.944 & 0.439 & 16.413 & 0.720 & 17.744 & 0.382 \\
    IBLA~\cite{7840002} & 15.929 & 0.710 & 17.531 & 0.291 & 18.914 & 0.710 & 13.379 & 0.319 & 16.953 & 0.721 & 16.476 & 0.331 \\
    Sea-thru~\cite{8954437} & 13.816 & 0.580 & 20.491 & 0.421 & 12.725 & 0.499 & 24.475 & 0.496 & 12.908 & 0.505 & 22.696 & 0.501 \\
    Bayesian-retinex~\cite{ZHUANG2021104171} & 18.749 & 0.829 & 13.391 & 0.262 & 15.640 & 0.669 & 18.679 & 0.369 & 17.586 & 0.747 & 15.341 & 0.321 \\
    UNTV~\cite{9548907} & 16.458 & 0.669 & 14.985 & 0.420 & 17.635 & 0.611 & 14.177 & 0.335 & 18.356 & 0.660 & 13.711 & 0.376 \\
    MLLE~\cite{9788535} & 18.740 & 0.814 & 11.482 & 0.234 & 15.142 & 0.633 & 16.089 & 0.323 & 17.868 & 0.730 & 12.782 & 0.278 \\
    HLRP~\cite{9854113} & 13.299 & 0.259 & 19.648 & 0.364 & 11.412 & 0.186 & 24.486 & 0.500 & 12.965 & 0.221 & 21.010 & 0.429 \\
    PCDE~\cite{10065491} & 15.830 & 0.701 & 15.901 & 0.335 & 14.451 & 0.586 & 17.823 & 0.395 & 15.751 & 0.647 & 15.761 & 0.379 \\
    ICSP~\cite{10167688} & 12.042 & 0.599 & 24.133 & 0.552 & 11.735 & 0.522 & 24.639 & 0.413 & 11.959 & 0.583 & 26.055 & 0.508 \\
    WWPF~\cite{10196309} & 18.596 & 0.822 & 11.458 & 0.218 & 15.954 & 0.648 & 15.261 & 0.337 & 17.902 & 0.739 & 12.708 & 0.283 \\
    ADPCC~\cite{zhou2023underwater} & 17.326 & 0.819 & 13.621 & 0.219 & 15.196 & 0.692 & 18.049 & 0.349 & 16.197 & 0.763 & 16.183 & 0.299\\
    \midrule
    UGAN~\cite{8460552} & 21.521 & 0.804 & 10.282 & 0.189 & 23.296 & 0.815 & 8.276  & 0.220 & 24.218 & 0.840 & 8.068  & 0.191 \\
    WaterNet~\cite{8917818} & \underline{22.82}  & \underline{0.907} & {\color{red}\textbf{8.518}}  & \underline{0.125} & 24.12  & 0.839 & 7.577  & 0.222 & 25.25  & \underline{0.877} & 7.433  & 0.164 \\
    FUnIE-GAN~\cite{9001231} & 19.167 & 0.800 & 12.604 & 0.217 & 22.624 & 0.720 & 8.254  & 0.230 & 21.660 & 0.744 & 9.901  & 0.240 \\
    UWCNN~\cite{LI2020107038} & 18.443 & 0.844 & 13.787 & 0.203 & 23.491 & 0.830 & 7.993  & 0.231 & 21.728 & 0.844 & 10.626 & 0.228 \\
    UColor~\cite{9426457} & 18.374 & 0.814 & 14.295 & 0.221 & 23.721 & 0.828 & 8.022  & 0.205 & 21.297 & 0.821 & 11.675 & 0.225 \\
    PUIE-Net~\cite{10.1007/978-3-031-19797-0_27} & 21.037 & 0.877 & 10.030 & 0.136 & 20.478 & 0.784 & 10.777 & 0.270 & 22.072 & 0.864 & 9.488  & 0.191 \\
    UIE-WD~\cite{9747781} & 20.275 & 0.848 & 13.159 & 0.198 & 17.795 & 0.760 & 14.086 & 0.292 & 19.233 & 0.803 & 14.735 & 0.284 \\
    TACL~\cite{9832540} & 19.831 & 0.761 & 11.297 & 0.222 & 20.990 & 0.782 & 11.314 & 0.213 & 22.972 & 0.828 & 9.731  & 0.176 \\
    GUPDM~\cite{10.1145/3581783.3612323} & 22.133 & 0.903 & 10.328 & 0.131 & 24.788 & \underline{0.847} & 7.416  & \underline{0.184} & \underline{25.325} & \underline{0.877} & 7.563  & \underline{0.150} \\
    U-Transformer~\cite{10129222} & 20.747 & 0.810 & 10.727 & 0.228 & \underline{24.992} & 0.829 & \underline{6.804}  & 0.238 & 25.151 & 0.838 & \underline{7.088}  & 0.221 \\
    PUGAN~\cite{10155564} & 20.524 & 0.812 & 11.144 & 0.216 & 22.580 & 0.820 & 8.746  & 0.212 & 23.135 & 0.836 & 8.989  & 0.216 \\
    CLUIE-Net~\cite{9965419} & 19.948 & 0.874 & 11.975 & 0.168 & 24.849 & 0.844 & 6.977  & 0.186 & 23.571 & 0.864 & 8.974  & 0.175 \\
    MBANet~\cite{XUE2023109041}  & 19.577 & 0.800 & 11.795 & 0.209 & 23.760 & 0.819 & 7.495  & 0.225 & 23.019 & 0.843 & 8.638  & 0.210 \\
    DeepWaveNet~\cite{10.1145/3511021} & 21.553 & 0.904 & 10.501 & 0.143 & 23.408 & 0.836 & 8.182  & 0.204 & 23.806 & 0.870 & 8.887  & 0.177\\
    Ours & {\color{red}\textbf{22.842}} & {\color{red}\textbf{0.923}} & \underline{9.098} & {\color{red}\textbf{0.108}} & {\color{red}\textbf{26.413}} & {\color{red}\textbf{0.868}}& {\color{red}\textbf{5.721}} & {\color{red}\textbf{0.130}} & {\color{red}\textbf{26.319}} & {\color{red}\textbf{0.892}} & {\color{red}\textbf{6.555}} & {\color{red}\textbf{0.116}} \\
    \bottomrule
    \end{tabular}
    \end{adjustbox}
    \label{table_comparison}
    
\end{table*}
\begin{table*}[ht]
    \centering
    \caption{Quantitative results of ablation experiments. \XSolidBrush and \CheckmarkBold respectively indicate the absence and presence of the module in the model. {\color[HTML]{FE0000} Red} and \textbf{bold} font indicates the best results.}
    \adjustbox{width=\linewidth}{
    \begin{tabular}{ccc|cccc|cccc|cccc}
    \toprule
    \multicolumn{3}{c|}{Module}
    & \multicolumn{4}{c|}{UIEB}
    & \multicolumn{4}{c|}{EUVP}
    & \multicolumn{4}{c}{LSUI}
    \\
    \midrule
    CPGB & NAGDM & ISF-Former
    & PSNR $\uparrow$ & SSIM $\uparrow$ & $\Delta$E $\downarrow$ & LPIPS $\downarrow$ 
    & PSNR $\uparrow$ & SSIM $\uparrow$ & $\Delta$E $\downarrow$ & LPIPS $\downarrow$
    & PSNR $\uparrow$ & SSIM $\uparrow$ & $\Delta$E $\downarrow$ & LPIPS $\downarrow$
    \\
    \midrule
    \XSolidBrush & \CheckmarkBold &\CheckmarkBold & 21.163 & 0.9 & 10.99 & 0.148 & 24.366 & 0.841 & 7.879 & 0.2 & 23.775 & 0.873 & 9.258 & 0.176 \\
    \CheckmarkBold & \XSolidBrush & \CheckmarkBold & 21.753 & 0.909 & 10.493 & 0.128 & 24.144 & 0.84 & 7.671 & 0.205 & 23.77 & 0.871 & 8.594 & 0.171 \\
    \CheckmarkBold &\CheckmarkBold & \XSolidBrush & 21.242 & 0.891 & 10.922 & 0.142 & 24.199 & 0.841 & 7.741 & 0.201 & 23.498 & 0.871 & 8.877 & 0.175\\    
    \XSolidBrush & \CheckmarkBold &\XSolidBrush & 20.462 & 0.883 & 11.647 & 0.155 & 24.16 & 0.84 & 8.011 & 0.199 & 23.087 & 0.867 & 9.769 & 0.18 \\    
    \XSolidBrush & \XSolidBrush &\CheckmarkBold & 20.441 & 0.89 & 11.653 & 0.152 & 24.114 & 0.838 & 8.084 & 0.203 & 23.219 & 0.868 & 9.052 & 0.176 \\
    \CheckmarkBold & \XSolidBrush & \XSolidBrush & 20.791 & 0.891 & 11.708 & 0.156 & 23.707 & 0.834 & 7.975 & 0.205 & 23.355 & 0.868 & 9.141 & 0.185 \\
    \CheckmarkBold & \CheckmarkBold & \CheckmarkBold & {\color[HTML]{FE0000} \textbf{22.842}} & {\color[HTML]{FE0000} \textbf{0.923}} & {\color[HTML]{FE0000} \textbf{9.098}} & {\color[HTML]{FE0000} \textbf{0.108}} & {\color[HTML]{FE0000} \textbf{26.413}} & {\color[HTML]{FE0000} \textbf{0.868}} & {\color[HTML]{FE0000} \textbf{5.721}} & {\color[HTML]{FE0000} \textbf{0.13}} & {\color[HTML]{FE0000} \textbf{26.319}} & {\color[HTML]{FE0000} \textbf{0.892}} & {\color[HTML]{FE0000} \textbf{6.555}} & {\color[HTML]{FE0000} \textbf{0.116}} \\
    \bottomrule
    \end{tabular}
    }
    \label{table:ablation}
\end{table*}

\subsection{Objective Function}
For model training, we use a linear combination of Mean Squared Error Loss $\mathcal{L}_{MSE}$ and Structural Similarity Index Loss $\mathcal{L}_{SSIM}$~\cite{1284395} as the objective function.

$\mathcal{L}_{MSE}$ focuses on pixel-level differences between the enhanced and target image. In the UIE task, using $\mathcal{L}_{MSE}$ encourages the enhanced image to have pixel values close to the target image. $\mathcal{L}_{MSE}$ is defined as:
\begin{equation}
    \mathcal{L}_{MSE}= \frac{1}{N} \sum_{i=1}^{N}(P_i-\hat{P_i})^2,
\end{equation}
where $P_i$ represents the value of $i$-th pixel in the target image and $\hat{P_i}$ represents the value of $i$-th pixel in the enhanced image, $N$ is the number of pixels in each image.

$\mathcal{L}_{SSIM}$ is designed to maximize the structural similarity between the enhanced and target image to improve the visual perception quality. $\mathcal{L}_{SSIM}$ is calculated as:
\begin{equation}
    \mathcal{L}_{SSIM}= \frac{(2\mu_{y} \mu_{\hat{x}} + C_1)(2\sigma_{y\hat{x}}+ C_2)}{(\mu_{y}^2+\mu_{\hat{x}}^2+C_1)(\sigma_{y}^2+\sigma_{\hat{x}}^2+C_2)},
\end{equation}
where $\mu_{y}$ and $\mu_{\hat{x}}$ represent the mean values of the target and enhanced images, respectively. $\sigma_{y\hat{x}}$ denotes the covariance between the two images, while $\sigma_{y}$ and $\sigma_{\hat{x}}$ represent their respective standard deviations. $C_1$ and $C_2$ are two constants to stabilize the division.

This combination strikes a balance between pixel-wise accuracy and perceptual quality. The objective function can be defined as:
\begin{equation}
    \mathcal{L}_{Total}=\mathcal{L}_{MSE}+\lambda * \mathcal{L}_{SSIM},
\end{equation}
where $\lambda$ is a tuneable hyperparameter. We empirically set it to 0.4.
\section{Experiments}

\subsection{Datasets and Evaluation Metrics}

\subsubsection{Datasets}
To comprehensively evaluate and enhance the performance of UIE, we utilized several underwater image datasets with a diverse range of real-world underwater environments and scenes, including Underwater Image Enhancement Benchmark (UIEB)~\cite{8917818}, Enhancing Underwater Visual Perception (EUVP)~\cite{9001231}, and Large Scale Underwater Image Dataset (LSUI)~\cite{10129222}. UIEB includes 950 real-world underwater images with varying degrees of degradation. Among them, 890 images have corresponding reference images (\ie input and target). EUVP is a vast underwater image dataset collected under various aquatic environments and light conditions. It includes images with reference and without reference of good and poor visual quality. LSUI is a large-scale underwater image collection with rich water scenes and types, which contains 4279 images and corresponding high-quality reference images.

\subsubsection{Evaluation Metrics}
In order to objectively evaluate the performance of our model, we employed several evaluation metrics in our experiments, including Peak Signal-to-Noise Ratio (PSNR), Structural Similarity Index Measure (SSIM)~\cite{1284395}, $\Delta$E~\cite{https://doi.org/10.1002/col.20070}, and Learned Perceptual Image Patch Similarity (LPIPS)~\cite{8578166}. PSNR  measures the quality of enhanced images relative to reference images. SSIM evaluates the variation in structural information, perceived illumination, and contrast between enhanced images and reference images. $\Delta$E evaluates color discrepancies between enhanced images and reference images. LPIPS assesses the similarity between two images in a way that reflects human perception.

\subsection{Implementation Details}
In our experiments, our model was implemented based on PyTorch and Linux and we trained the model on 4 GeForce RTX 2080Ti GPUs. During the training process, we set the training epoch to 300 and the batch size to 4. For model optimization, we selected the AdamW optimizer with parameters $\beta_1 = 0.9$, $\beta_2 = 0.999$, and an initial learning rate of $2\times10^{-4}$. Furthermore, to adjust the learning rate dynamically, we adopted the Cosine Annealing Learning Rate Scheduler (CosineAnnealingLR). It reduces the risk of getting stuck in local minima by effectively ``warming up'' the learning rate periodically. We set the learning rate range between $1\times10^{-6}$ and $2\times10^{-4}$. 

For training, we randomly selected 800 images from UIEB and 2000 images each from EUVP and LSUI as the training set. EUVP includes data categorized into three groups: Dark, ImageNet, and Scenes. From each category, we randomly selected 800, 700, and 500 images, respectively. Images were resized to $256\times256$ to facilitate the model training. To enhance data diversity, we applied various augmentation techniques such as random flipping, rotation, transposition, mixup, and cropping to the images. It is important to note that only images with references were used during the training process.

For testing, we selected the remaining 90 images from UIEB, along with 200 randomly chosen images each from EUVP and LSUI, to form the testing set. The images in the testing set do not overlap with those in the training set.
\subsection{Comparisons with State-of-the-arts}
To comprehensively evaluate the performance of our model, we compared it with 25 state-of-the-art methods. All the learning-based models were retrained using the same training parameters on a training set split in the same way.

\subsubsection{Quantitative Results}
Table~\ref{table_comparison} displays the quantitative results of comparative experiments between our model and other state-of-the-art methods across three datasets. The table is divided into two sections: the upper focuses on traditional methodologies and the lower on learning-based approaches. As shown in the table, compared to traditional methods, our model exhibits significantly superior performance on three datasets. When compared to learning-based methods, on the UIEB dataset, our model slightly lags behind WaterNet~\cite{8917818} in the $\Delta$E metric but has achieved the best performance in the other three metrics. On the EUVP and LSUI datasets, our model surpasses the others across all four metrics, securing the best performance. Overall, our model demonstrates excellent performance across various datasets, reflecting its adaptability to different aquatic environments.

\subsubsection{Qualitative Results}
In order to qualitatively demonstrate the superiority of our model over other methods, we provide some results of visual comparison. Fig.~\ref{fig:compare_tradition} depicts our model's performance in comparison to several benchmark traditional methods. It is evident that our model demonstrates the best visual performance. Owing to inherent limitations, traditional methods are only applicable to images captured in specific aquatic environments and light conditions. They may not effectively remove haze or might introduce unrealistic colors in the enhancement of images captured in different water environments, lacking universality. Fig.~\ref{fig:compare_learningbased} shows the visual comparison of our model with some representative learning-based methods. Compared to traditional methods, learning-based approaches adapt better to images taken in various underwater environments, demonstrating stronger adaptive capabilities. Moreover, in comparison with other deep learning methods, our model not only recovers the texture and details of the images but also enhances the color saturation and contrast more effectively, making the improved images more similar to the target images.

\subsection{Ablation Studies}
In order to demonstrate the efficacy of the module we proposed for UIE, we conducted ablation experiments on three datasets: UIEB, EUVP, and LSUI, with the experimental results shown in Table~\ref{table:ablation}. During the experiments, we tested three modules: CPGB, NAGDM, and ISF-Former. By removing one or two of them, the experimental results indicated that our complete model surpassed other models with modules ablated, demonstrating the best performance. This proves the importance of the proposed modules for UIE.

\begin{figure}[ht]
\centering
    \begin{minipage}[b]{\linewidth}
    \centering
        \begin{minipage}[b]{0.32\linewidth}
            \centering
            \centerline{\includegraphics[width=\linewidth]{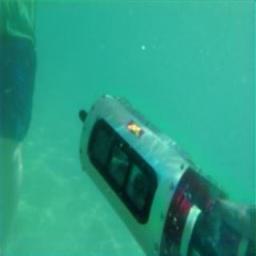}}            \centerline{(a)}\medskip
        \end{minipage}
        \hfill
        \begin{minipage}[b]{0.32\linewidth}
            \centering
            \centerline{\includegraphics[width=\linewidth]{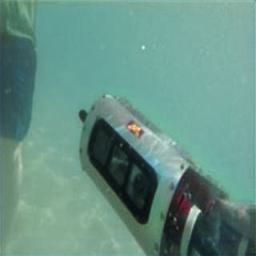}}            \centerline{(b)}\medskip
        \end{minipage}
        \hfill
        \begin{minipage}[b]{0.32\linewidth}
            \centering
            \centerline{\includegraphics[width=\linewidth]{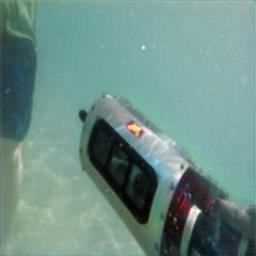}}            \centerline{(c)}\medskip
        \end{minipage}
        \hfill
    \end{minipage}

    \begin{minipage}[b]{\linewidth}
    \centering
        \begin{minipage}[b]{0.32\linewidth}
            \centering
            \centerline{\includegraphics[width=\linewidth]{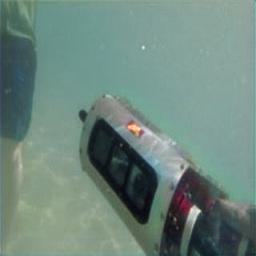}}            
            \centerline{(d)}\medskip
        \end{minipage}
        \hfill
        \begin{minipage}[b]{0.32\linewidth}
            \centering
            \centerline{\includegraphics[width=\linewidth]{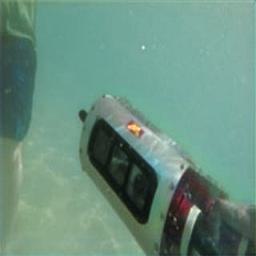}}            
            \centerline{(e)}\medskip
        \end{minipage}
        \hfill
        \begin{minipage}[b]{0.32\linewidth}
            \centering
            \centerline{\includegraphics[width=\linewidth]{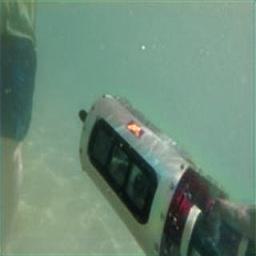}}            \centerline{(f)}\medskip
        \end{minipage}
        \hfill
    \end{minipage}

    \begin{minipage}[b]{\linewidth}
    \centering
        \begin{minipage}[b]{0.32\linewidth}
            \centering
            \centerline{\includegraphics[width=\linewidth]{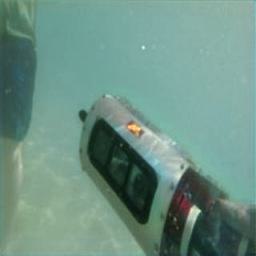}}            \centerline{(g)}\medskip
        \end{minipage}
        \hfill
        \begin{minipage}[b]{0.32\linewidth}
            \centering
            \centerline{\includegraphics[width=\linewidth]{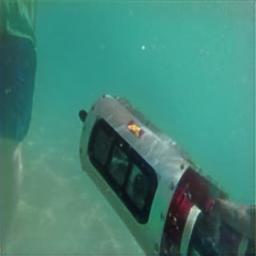}}            \centerline{(h)}\medskip
        \end{minipage}
        \hfill
        \begin{minipage}[b]{0.32\linewidth}
            \centering
            \centerline{\includegraphics[width=\linewidth]{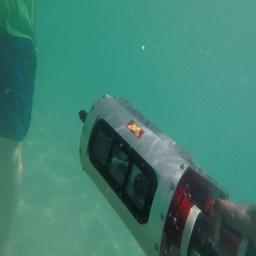}}            \centerline{(i)}\medskip
        \end{minipage}
        \hfill
    \end{minipage}
    \caption{Ablation study on the effectiveness of different components in the proposed model. (a) Input underwater image. (b) Result w/o CPGM. (c) Result w/o NAGDM. (d) Result w/o ISF-Former. (e) Result w/o CPGM and ISF-Former. (f) Result w/o CPGM and NAGDM. (g) Result w/o ISF-Former and NAGDM. (h) Result of the full proposed model. (i) Target image. The full model achieves the best visual quality, demonstrating the importance of each component.}
    \label{fig:ablation}
    
\end{figure}

\subsection{User Study}
To evaluate the subjective visual quality of enhanced underwater images, we conducted a comprehensive user study. This study aimed to assess the perceptual quality of images enhanced by our proposed method compared to state-of-the-art techniques and the original degraded images.

\subsubsection{Study Design}
We selected a diverse set of 50 underwater images from our test dataset, covering various underwater scenes and degradation levels. These images were enhanced using our proposed method, as well as three state-of-the-art techniques: HLRP~\cite{9854113}, CLUIE-Net~\cite{9965419}, and PUGAN~\cite{10155564}. The original degraded images were also included in the study.

A total of 30 participants were recruited for the study. The participants included both experts in image processing (10) and non-experts (20) to ensure a balanced evaluation. All participants reported normal or corrected-to-normal vision and were naive to the purpose of the experiment.

\subsubsection{Procedure}
The enhanced images were presented to participants in a randomized order on calibrated displays in a controlled environment with consistent lighting. Participants were asked to rate each image on a 5-point Likert scale: (1) Poor, (2) Fair, (3) Good, (4) Very Good, (5) Excellent. Participants were instructed to consider overall image quality, including color vividness, contrast, detail preservation, and naturalness. They were given unlimited time to view and rate each image.

We calculated the Mean Opinion Score (MOS) for each method by averaging the scores across all participants and images. The results are presented in Table~\ref{tab:mos}.

\begin{table}[ht]
\centering
\caption{MOS Results. Scores are presented as mean $\pm$ standard deviation.}
\begin{tabular}{cc}
\toprule
Method    & MOS \\ \midrule
Input     & $2.14\pm0.62$               \\
HLRP~\cite{9854113}      & $3.26\pm0.58$               \\
CLUIE-Net~\cite{9965419} & $3.82\pm0.45$               \\
PUGAN~\cite{10155564}     & $3.95\pm0.41$               \\
Ours      & $4.31\pm0.37$               \\ \bottomrule
\end{tabular}
\label{tab:mos}
\end{table}

As shown in Table~\ref{tab:mos}, our proposed method achieved the highest MOS of 4.31, outperforming other state-of-the-art methods and the original input images. This indicates that participants consistently perceived our enhanced images as having superior visual quality.

The results of our user study strongly support the effectiveness of our proposed method in enhancing underwater images. Participants consistently rated images enhanced by our method higher than those produced by other state-of-the-art techniques. These subjective results complement our quantitative evaluations, providing a comprehensive assessment of our method's performance. The high MOS achieved by our method suggests that it not only improves objective image quality metrics but also enhances the visual appeal and perceptual quality of underwater images, which is crucial for many practical applications.

\section{Conclusion}
In this paper, we propose a novel deep unfolding network for underwater image enhancement that integrates color priors and inter-stage feature transformation. The proposed model incorporates three key components: a Color Prior Guidance Block, a Nonlinear Activation Gradient Descent Module, and an Inter Stage Feature Transformer. Experimental results on multiple underwater image datasets demonstrate the superiority of the proposed model over state-of-the-art methods in both quantitative and qualitative evaluations. The proposed model offers a promising solution for underwater image enhancement, enabling more accurate and reliable scientific analysis in various marine applications. Future work may focus on further improving the model's efficiency and adaptability to diverse underwater environments and exploring its potential in other low-light image enhancement tasks.

\section*{Acknowledgment}
This work was supported by the Science and Technology Development Fund, Macau SAR, under Grant 0141/2023/RIA2 and 0193/2023/RIA3.

\bibliographystyle{IEEEtran}
\bibliography{ref}

\end{document}